# LTNER: Large Language Model Tagging for Named Entity Recognition with Contextualized Entity Marking

Faren Yan[1,2], Peng Yu[1,2], Xin Chen[1,2] (✉)

[1] Computer Network Information Center, Chinese Academy of Sciences, Beijing, China
[2] University of Chinese Academy of Sciences, Beijing, China
`fryan@cnic.cn,yupeng23@mails.ucas.ac.cn,chx@cnic.cn`

**Abstract.** The use of LLMs for natural language processing has become a popular trend in the past two years, driven by their formidable capacity for context comprehension and learning, which has inspired a wave of research from academics and industry professionals. However, for certain NLP tasks, such as NER, the performance of LLMs still falls short when compared to supervised learning methods. In our research, we developed a NER processing framework called LTNER[1] that incorporates a revolutionary Contextualized Entity Marking Gen Method. By leveraging the cost-effective GPT-3.5 coupled with context learning that does not require additional training, we significantly improved the accuracy of LLMs in handling NER tasks. The F1 score on the CoNLL03 dataset increased from the initial 85.9% to 91.9%, approaching the performance of supervised fine-tuning. This outcome has led to a deeper understanding of the potential of LLMs.

## 1 Introduction

Named Entity Recognition (NER) is an essential component of Natural Language Processing (NLP), playing a vital role in various fields such as text information extraction, information retrieval, and construction of knowledge graphs. The objective of NER is to identify and classify entities in a given text into predefined categories. Although the advent of Transformer models [1] such as Bert [2] and T5 [3] has considerably enhanced the processing capabilities of neural networks on certain datasets, the practical application cost is relatively high due to the substantial data annotation and model training required.

Recently, Large Language Models (LLMs) have emerged as a transformative force in the field of NLP. GPT-3.5 serves as an example of these models' significant abilities in understanding and generating context-rich text. Owing to their numerous params and extensively applied training data [4-5], these models demonstrate excep-tional capacity for capturing contextual information. Moreover, the convenience of using these models is notable, as they can be easily directed through Prompts to complete an array of tasks, greatly accelerating the deployment of various NLP applications. However, a significant gap exists between LLMs and traditional fine-tuning methods when it comes to NER tasks due to fundamental differences in their output modalities: while NER emphasizes precise annotation [6], GPT models focus on generation; moreover, the issue of hallucinations [7] further compounds the challenge.

---
[1] Code is available at https://github.com/YFR718/LTNER

In this paper, we introduce LTNER, a method employing contextual marking to leverage the Context-Learning abilities of GPT-3.5 for the improvement of NER tasks without the need for fine-tuning or large-scale training. Experiments conducted on multiple datasets show that our method significantly outperforms existing context NER techniques and closely matches the accuracy of traditional supervised learning methods.

In sum, our contributions are as follows:

1. **An innovative NER method.** We have proposed a simple yet effective Contextual Entity Marking Generation method for large language models (LTNER), which achieves an accuracy close to that of mainstream supervised learning without the necessity for training or fine-tuning.
2. **Robustness of the model.** We have provided ample empirical evidence of LTNER's excellent performance with few contextual examples, limited labeled data, and low cost.
3. **Techniques for optimizing prompt engineering.** We have explored various techniques for optimizing prompt engineering, such as prompt format and role designation, offering valuable insights for future research.

## 2 LTNER

This chapter provides a detailed exposition of the design principles behind LTNER. Section 2.1 defines the NER task, Section 2.2 elaborates on the design of the token generation learning method, and Section 2.3 describes the operational flow of the overall system architecture.

### 2.1 Definition of the NER Task

NER is fundamentally a sequence labeling problem: each token within the text must be assigned a label, indicating whether and how it constitutes a specified entity category. Given a text sequence $X = (x_1, x_2, \ldots, x_n)$, where $x_i$ represents the i-th token in the sequence, the objective of the NER task is to produce a set of entity labels $Y = (y_1, y_2, \ldots, y_n)$, with each $y_i$ selected from a predetermined label set. This label set typically includes $B - Person$ (indicating the beginning of a person's name), $I - Person$ (the continuation of a person name entity), $B - Organization$ (the beginning of an organization name), $I - Organization$ (the continuation of an organization name entity), and other categories such as locations, dates, etc. The $O$ label represents a non-entity.

In traditional supervised learning approaches, a NER model is defined as a conditional probability distribution $P(Y \mid X)$ of the entity label sequence given the text sequence. The training objective of the model is to maximize this conditional probability, which is usually achieved by maximizing the log-likelihood function:

$$\max_{\theta} \sum_{i=1}^{n} logP(y_i \mid X, y_1, y_2, \ldots, y_{i-1}; \theta) \tag{1}$$

Here, $\theta$ represents model parameters, and $y_1, y_2, \ldots, y_{i-1}$ denote the labels for the first i-1 tokens in the sequence. In this way, the NER model learns to allocate the most suitable label for each token, based on the context.

### 2.2 Contextualized Entity Marking Gen Method

For generative models like GPT, given a prompt sequence $P = (p_1, p_2, \ldots, p_m)$, the goal shifts to maximizing the conditional probability of the desired entity label sequence $Y$ given the text sequence $X$ and the prompt sequence $P$. This can be expressed using Bayes' theorem as follows:

$$P(Y|X,P;\theta) = \frac{P(Y,X,P|\theta)}{P(X,P;\theta)} \tag{2}$$

where $P(Y,X,P|\theta)$ represents the joint probability of the text sequence $X$, the prompt sequence $P$, and the entity label sequence $Y$ under the model parameters $\theta$, and $P(X,P;\theta)$ is the marginal probability of the text sequence $X$ and the prompt sequence $P$, serving as a normalizing factor to ensure the sum of the probabilities equals 1.

To enhance model performance without altering the model parameters, the primary strategies include:

**Designing a Simplified Output Format.** To minimize the differences between NER annotation tasks and GPT generation tasks, we have devised a special label marking mechanism. Entities' start and end positions are marked with '##', and the word following '##' represents the entity category. Text not enclosed by '##' does not belong to any entity. This annotation method effectively completes the NER marking mapping while also aligning with GPT's generative pattern, allowing the model to reproduce the original text and insert minimal labels in appropriate places, significantly reducing the interference of extraneous information.

**Providing Richer and Utility Context Information.** Context learning is a common method to stimulate the learning capability of large models. When a certain amount of annotated data is available, leveraging vector-based retrieval to obtain the most relevant context as the input example and using the annotated results as the output, significantly enhances the large model's ability to perform such tasks.

## 2.3 Operational Framework

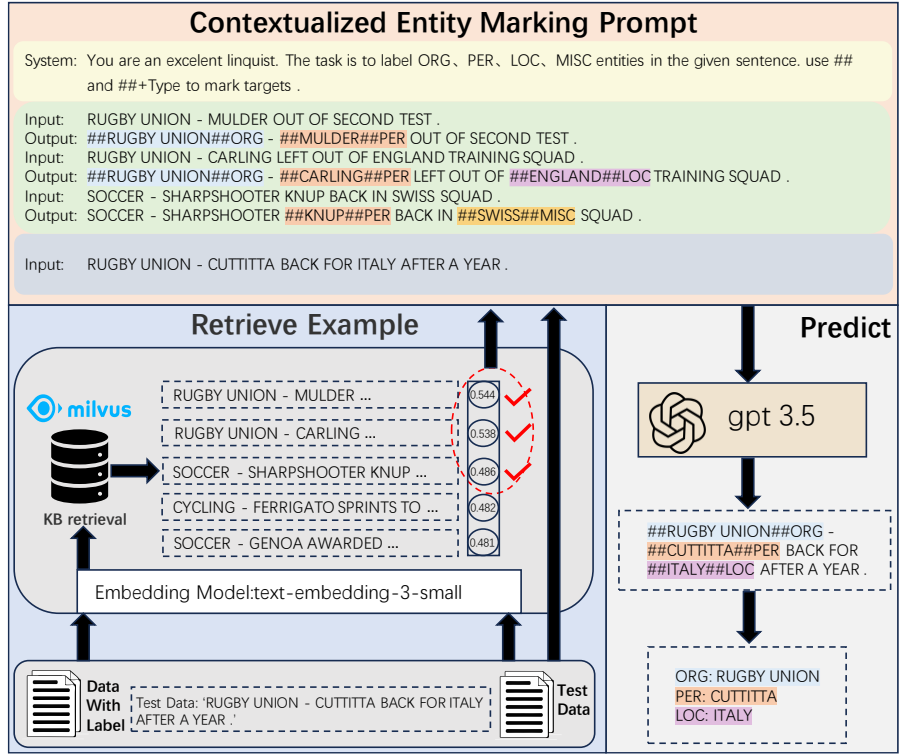

**Fig. 1.** The LTNER system framework, comprising three parts: data encoding and vector storage, vector retrieval to construct contextual entity marker learning examples, and result generation and parsing.

Our system operation is divided into three stages, as shown in Fig. 1. Firstly, we establish a knowledge base by vectorizing the text of training data; vectors, original texts, and annotation results are jointly stored in the vector database. Subsequently, for each piece of test data, the same encoding process is performed to generate the vector, and with these vectors, we retrieve the most similar N pieces of data from the knowledge base. These data are processed according to the previously described annotation method to form context learning examples. Finally, the sentences to be annotated are concatenated with these examples, and the large-scale model generates output results. By parsing these results according to the annotation rules, we derive the predicted output for the NER task.

## 3 Experiment

### 3.1 Setup

**Datasets.** For the empirical data of the NER task, we selected the CoNLL-2003 dataset[8], which is extensively employed in experimental research. The dataset is available in both English and German versions, and for this study, we utilized the English version. It encompasses four categories of entities: person (PER), location (LOC), organizations (ORG), and miscellaneous (MISC). The dataset contains over 20,000 sentences and more than 3,500 entities. To further showcase the method's capacity for generalization, we incorporated an additional dataset, WNUT 2017 [9]. This dataset encompasses six distinct entity types and presents a notable challenge due to its text rife with diverse noise patterns. Its intricate nature renders identification more arduous, rendering it an apt choice for evaluating the efficacy of our approach in managing noisy textual data.

In the data preprocessing phase, we first converted the data from Inside-Outside-Beginning (IOB) format to an easily manipulable raw-text-to-label Json format for subsequent use in Prompt construction.

```
AL-AIN NNP B-NP B-
LOC
, , O O
United NNP B-NP B-
LOC
Arab NNP I-NP I-LOC
Emirates NNPS I-NP
I-LOC
1996-12-06 CD I-NP O
```

IBO format

```
"sentence": "AL-AIN ,
United Arab Emirates 1996-
12-06",
"label": {
    "LOC": [
        "AL-AIN",
        "United Arab Emir-
ates "
    ]
}
```

Json fomat

### 3.2 Main Results

Our experiment aims to evaluate the effectiveness of contextual learning in the NER task using LTNER. Table 1 presents a comparison of various methods including ours. Notably, the supervised learning and model fine-tuning are represented by InstructUIE[10] and GPT-NER[6]. InstructUIE uses instruction fine-tuning to infuse knowledge, whereas GPT-NER enhances adaptability to NER tasks by training a specialized NER vector encoder. The methods discussed in the middle section are those that do not require fine-tuning, such as CodeIE[11] and Code4UIE[12], which employ a code format output to tightly integrate NER tasks with the powerful code generation capabilities of large models. This class of methods, using content generative models, has shown to be more effective than those using conversational models.

The no-fine-tuning version of GPT-NER uses a conventional encoding model to process sentences, outputting labels. However, it extracts only one type of entity at a time and introduces optimization steps like the secondary inspection by large models. The penultimate method employs a Json format output that is easy to parse and well-structured, widely adopted by many mainstream applications today. Finally, our research achieves results through a synthesis of label annotation output and role optimization.

**Table 1.** Documents the comparison of LTNER with other mainstream methods, including the model name, learning approaches, backbone networks, and the test results of precision, recall, and F1 score.

| Method | Paradigm | Backbone | CoNLL 2003 | | | WNUT 2017 |
|---|---|---|---|---|---|---|
| | | | P(%) | R(%) | F1(%) | F1(%) |
| GoLLIE | SFT | Code-LLaMA 34B | - | - | **93.10** | 54.30 |
| InstructUIE | SFT | Flan-T5-11B | - | - | 92.94 | - |
| GPT-NER | SFT+ICL | Text-davinci-003 | 89.76 | 92.06 | 90.91 | - |
| GPT-NER | ICL | Text-davinci-003 | 83.73 | 88.07 | 85.90 | 40.51 |
| CodeIE | ICL | Code-davinci-002 | - | - | 82.32 | 39.67 |
| Code4UIE | ICL | Text-davinci-003 | - | - | 83.60 | 41.94 |
| Json | ICL | GPT-3.5-turbo | 86.50 | 83.57 | 85.01 | 40.88 |
| LTNER(ours) | ICL | GPT-3.5-turbo | 92.86 | 90.98 | **91.91** | 49.74 |

From the data on the right side of the table, it's apparent that prior methods employing ICL average around an 85% F1 score, while those that undergo model fine-tuning hover around 93%. Our context-aware label formatted output method achieves an F1 score of 91.9%, significantly surpassing the existing ICL techniques and closely approaching the fine-tuned models (only a 1% gap). For the WNUT dataset, although the performance gap between LTNER and the fine-tuned model widened (4.6%), LTNER still maintains a significant lead of around 8% over the commonly used ICL method. These findings indicate that our method possesses significant competitive advantages in ICL scenarios.

### 3.3 Ablation experiment

**Table 2.** Records the comparison results of the ablation study, which include variables such as the generation mode, format of the labels, and the role settings for contextual learning. Constants include the number of context samples, the vector retrieval method, and the underlying model, among others.

| Pattern | Shots | Tag Combinations | Role Setting | P(%) | R(%) | F1(%) |
|---|---|---|---|---|---|---|
| Json | 30 | - | SUA | 84.61 | 82.96 | 83.78 |
| Tag | 30 | @@Entry##Label | SUA | 91.66 | 88.33 | 89.97 |
| Tag | 30 | ##Entry##Label | SUA | 91.42 | 89.48 | 90.44 |
| Tag | 30 | @@Entry##Label | AAA | 91.53 | 89.23 | 90.37 |
| Tag | 30 | ##Entry##Label | AAA | 91.42 | 90.54 | 90.98 |

In this section of the ablation study, we meticulously investigated the impact of factors such as generation mode, label formatting, and the role setting in context learning on the performance of the LTNER model. To ensure the accuracy and comparability of our experiments, parameters such as the number of context samples, vector retrieval methods, and the underlying model were held constant , using the CoNLL03 dataset.

In comparing generation modes, standard Json format output was juxtaposed with label-based output. With the same 30-shot setting, switching to label-based output resulted in a significant improvement in the F1 score, from 83.7% to 89.9%, indicating a notable enhancement in model performance. Further analysis of various label formats revealed that using '##' as both the beginning and ending symbols for tags, despite a modest decrease in precision (P value), markedly improved the recall rate (R value), and consequently the F1 score. This label format aligns more closely with the training data conventions of LLMs and is better adhered to by the model. For the effects of over ten other label formats, please see Appendix I.

For dialogue-based models, assigning appropriate roles when invoking the API is crucial. Typically, we designate the background information as the 'System' role, placed at the beginning of the conversation, and the context examples' queries and answers as the 'User' and 'Assistant' roles respectively, with the text to be identified also portrayed by the 'User' role. This method is referred to as the SUA mode. Experiments have demonstrated that setting all roles to 'Assistant' (thus the AAA mode) also aids in enhancing the recall rate. Such role configurations better suit the nature of NER tasks, given that their inputs and outputs are fixed text and tags, unlike those in a user-machine dialogue setting. For additional role combination effects, see Appendix II.

Ultimately, by integrating the double-pound tag notation with the AAA role control strategy, an approximate 1% increment in the F1 score was achieved compared to using the standard label format; as opposed to the conventional Json format output, we observed a significant 7% increase.

The findings underscore the superiority of our approach and provide high-quality empirical evidence for standards set by top-tier conferences in the field of computer science.

## 4 Analysis

In this section, we will delve into an in-depth analysis of several key factors that affect the performance of the LTNER model. This includes the number of different contextual examples, the quantity of labeled data, and the performance under various levels of expenditure, further exploring the robustness of the LTNER and its advantages in terms of low cost. We conducted these analyses using the CoNLL dataset.

### 4.1 Diffrent Number of Contextual Example

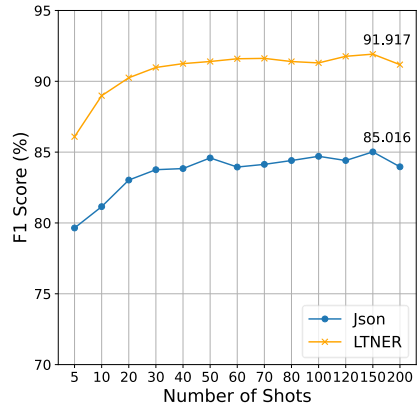 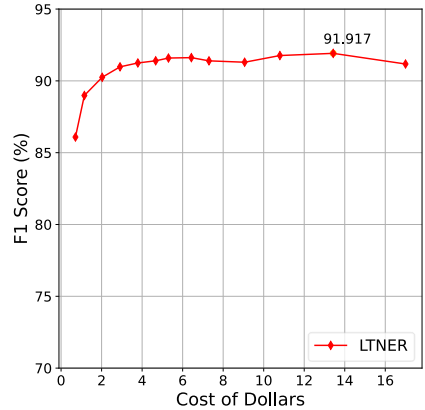

**Fig. 2.** The Relationship between the Number of Contextual Examples and F1 Score

**Fig. 3.** The Relationship between F1 Scores and Monetary Expenditure

Contextual learning has been proven to be an extremely effective method for improving the performance of large models [13]. The contextual examples of the LTNER are presented based on the marked label format. As shown in Fig. 2, the relationship between the number of examples and the accuracy of the model is apparent. From the figure, we can observe that, starting from 5 samples, the accuracy of the model begins to increase significantly and gradually stabilizes when the number of samples is between 50 and 70. At this point, the accuracies of the Json output format and the LTNER model are 91.61% and 84.59%, respectively. When the number of samples increases to 150, the accuracy of the LTNER model reaches its highest value of 85.08%, surpassing the Json output format, with an accuracy improvement of nearly 7%. As the number of samples continues to rise, there is a slight decline in accuracy, which may be due to the phenomenon of long-context forgetting

### 4.2 Diffrent Monetary Expenditure

Currently, large model deployment is more convenient than any deep learning technology before, mainly due to its powerful context comprehension capabilities, eliminating the need for extensive data annotation and model training, as well as separate deployment. The infrastructure for large models is now very mature, exemplified by OpenAI's gpt-3.5-turbo, which has become cost-effective and responsive following several iterations, with a cost of only $0.5 per million input tokens. This section aims to control the number of input context examples and test the accuracy distribution of the dataset at varying costs.

As observed from the results in Fig. 3, there is a significant increase in F1 scores within the cost interval of $0 to $5. With an expenditure of $3, the accuracy reaches 91%, indicating extremely cost-effective performance. At about $5, the F1 score reaches a turning point at 91.62%. Subsequent improvements in accuracy tend to plateau as the expenditure increases, consistent with findings from previous sample size experiments. These results highlight the low-cost advantage of LTNER.

## 4.3 Diffrent Number of Annotated Data

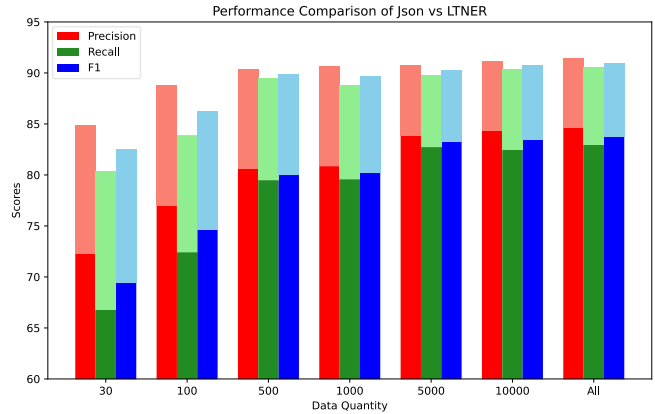

**Fig. 4.** The Relationship between the Number of Annotated Data and F1 Score (With the Number of Contextual Examples Fixed at 30)

Another significant advantage of employing large language models for NLP tasks is that they do not rely on the vast amounts of annotated data required by traditional supervised learning. A handful of annotated examples can yield quite impressive outcomes. This section of the study examines the results of NER tasks under different volumes of annotated data. To control for variables, we fixed the number of contextual examples at 30 and built a vector retrieval library by randomly selecting a certain amount of annotated data.

The results depicted in Fig. 4 demonstrate that with just 30 annotated examples, the F1 scores for Json and LTNER can reach 79.75% (69.43/83.72) and 90.75% (82.56/90.97) respectively, of the performance attained with full data annotation. Notably, LTNER exhibits markedly superior learning efficiency with the same data amount. When the volume of annotated data increases from 30 to 500, the F1 score rises rapidly, with LTNER achieving 98.7% (89.86/90.97) of the performance of full data annotation with only 1/30 of the annotated data. However, as the volume of data continues to expand, the rate of performance improvement slows down. This experiment vividly illustrates the potent learning capabilities of large models when presented with a small number of samples; a modest amount of data annotation can lead to such high output efficacy.

## 5    Related Work

**Generative Named Entity Recognition.** In the field of NLP, NER has become an increasingly studied direction. This method enhances the accuracy of entity recognition by controlling the format of the model's output text and performing multi-step inference. For instance, CodeIE and Code4UIE, leveraging the powerful code generation capabilities of large models, transform the entity recognition task into a code content generation task, thus achieving better structured output. GPT-NER employs a concise tagging and self-checking mechanism, concentrating on extracting entities of a single category. Our research reveals that named entity recognition fundamentally requires strong context support. Spreading out the entity recognition process might lead to misclassifications of the same entity into multiple categories due to a lack of a coherent context.

**Multi-stage Named Entity Recognition.** Decomposing complex tasks into multiple steps and solving them individually is considered an effective way to enhance the performance of LLMs. chatIE [14] is a prime example; it deconstructs the information extraction task into a multi-turn dialogue process: first identifying the types of entities to be recognized, then precisely extracting them. The research [15] proposes that by separating content generation from the process of structuring, the model can concentrate on each step independently, thus alleviating the pressure of handling two orthogonal tasks simultaneously. The quality of task prompts is crucial to the model's performance; the zero-sample self-annotation plus checking approach [16] enables the acquisition of a knowledge base in an unsupervised manner, followed by enhancing the inference based on these self-annotated examples. C-ICL [17] utilizes the construction of positive and negative examples for contextual learning demonstrations, thereby strengthening the LLMs' ability to extract entities and relationships.

**Fine-tuned Named Entity Recognition.** In the realm of fine-tuned named entity recognition, GPT-NER has developed an entity vector encoder to improve the efficiency and effectiveness of retrieving examples. GoLLIE [18] explores fine-tuning LLMs to meet precise instructional requirements, thus enhancing the zero-shot performance of LLMs in unseen information extraction (IE) tasks. InstructUIE utilizes structured instructions to fine-tune LLMs, which boosts UIE's ability to consistently simulate different IE tasks and capture the dependencies between them. Meanwhile, PaDeLLM-NER [19] significantly accelerates the reasoning speed for NER tasks by parallel decoding all mentions. The study [20] also finds that including negative examples in the training process can significantly improve the model's recognition performance for various tasks.

## 6    Conclusion

In this study, we introduce an innovative context marking extraction method named LTNER. By adopting a simple tagging generation format, this method substantially enhances the performance of large-scale language models on NER tasks. The outcomes not only surpass existing context-learning based methods but also approach the effects of model fine-tuning. Significantly, our experimental results demonstrate that

LTNER achieves effective entity extraction with few samples, sparse annotation data, and at a low cost, thereby offering novel approaches for the rapid deployment and application of NER tasks. We believe that potential future research directions include exploring the performance potential of different large-scale models such as GPT-4, improving LTNER to enhance its recognition capabilities in scenarios involving nested entities, and integrating the automated entity extraction capabilities of large-scale models with a broader range of practical applications to promote the extensive application of NLP technologies.

## A. Appendix

**Analyzing the Relationship Between Different Tags and Accuracy**

To rapidly ascertain the distinctions among various tags, we conducted experiments using the first 500 entries of the test set, with the number of context examples set at 30.

**Table 3.** The Relationship Between Tags and NER Performance

| Tag Combinations | P(%) | R(%) | F1(%) |
| --- | --- | --- | --- |
| ["##", "@@"] | 91.71 | 86.58 | 89.07 |
| ["@@", "##"] | 91.19 | 86.28 | 88.66 |
| ["@@", "##"] | 91.72 | 86.69 | 89.13 |
| ["##", "##"] | 91.44 | 89.02 | 90.22 |
| ["@@", "@@"] | 89.40 | 88.40 | 88.90 |
| ["@", "#"] | 91.92 | 82.20 | 86.79 |
| ["@", "@"] | 91.43 | 87.89 | 89.63 |
| ["#", "#"] | 91.23 | 86.78 | 89.47 |
| ["#", "@"] | 92.38 | 85.15 | 88.62 |
| ["[", "]"] | 0 | 0 | 0 |
| ["", "", ""] | 0 | 0 | 0 |
| ["<", ">"] | 89.11 | 88.20 | 88.65 |
| ["(", ")"] | 0 | 0 | 0 |
| ["+", "+"] | 0 | 0 | 0 |
| ["?", "?"] | 0 | 0 | 0 |
| ["%", "%"] | 88.31 | 87.59 | 87.95 |
| ["'", "'"] | 86.93 | 85.34 | 86.13 |
| ["{", "}"] | 91.40 | 88.71 | 90.04 |
| ["{{", "}}"] | 91.12 | 88.71 | 89.90 |
| ["[[", "]]"] | 0 | 0 | 0 |
| ["''", "''"] | 0 | 0 | 0 |
| ["<<", ">>"] | 91.67 | 89.52 | 90.58 |
| ["((", "))"] | 0 | 0 | 0 |
| ["%%", "%%"] | 89.39 | 89.12 | 89.26 |
| ["", ""] | 90.34 | 89.42 | 89.87 |

**Analyzing the Relationship Between Different Roles and Accuracy**

To rapidly ascertain the distinctions among different roles, we conducted experiments using the entire set of test data, with the number of context examples set at 30.

Table 4. The Relationship Between Role Settings and NER Performance

| Role Setting | P(%) | R(%) | F1(%) |
|---|---|---|---|
| UUU | 75.13 | 82.98 | 78.86 |
| AAA | 85.49 | 84.31 | 84.90 |
| SUU | 60.77 | 81.57 | 69.65 |
| SAA | 85.04 | 83.37 | 84.19 |
| SUA | 84.84 | 83.36 | 84.10 |
| AUA | 85.46 | 83.18 | 84.30 |
| UUA | 85.00 | 83.53 | 84.26 |